\documentclass[11pt,a4paper]{article}
\usepackage[hyperref]{acl2020}
\usepackage{times}
\usepackage{latexsym}

\usepackage[utf8]{inputenc}
\usepackage{graphicx}
\usepackage{lipsum}
\usepackage[justification=centering]{caption}
\usepackage{tabularx, multirow, booktabs}
\usepackage{wrapfig}
\usepackage{microtype}
\usepackage[utf8]{inputenc}

\aclfinalcopy % Uncomment this line for the final submission
\title{Emotions are Subtle: Learning Sentiment Based Text Representations Using Contrastive Learning}

\author{Ipsita Mohanty \\
  Carnegie Mellon University \\
  Pittsburgh, PA \\
  USA \\
  \textmd{imohanty@alumni.cmu.edu} \\\And
  Ankit Goyal \\
  Carnegie Mellon University \\
  Pittsburgh, PA \\
  USA \\
  \textmd{agoyal3@alumni.cmu.edu} \\\And
  Alex Dotterweich \\
  University of California \\
  Berkeley, CA \\
  USA \\
  \textmd{alexdotterweich@berkeley.edu} \\}

\date{}

\begin{document}
\maketitle
\begin{abstract}
Contrastive learning techniques have been widely used in the field of computer vision as a means of augmenting datasets. In this paper, we extend the use of these contrastive learning embeddings to sentiment analysis tasks and demonstrate that fine-tuning on these embeddings provides an improvement over fine-tuning on BERT-based embeddings to achieve higher benchmarks on the task of sentiment analysis when evaluated on the DynaSent dataset. We also explore how our fine-tuned models perform on cross-domain benchmark datasets. Additionally, we explore upsampling techniques to achieve a more balanced class distribution to make further improvements on our benchmark tasks.
\end{abstract}

\section{Introduction}
Sentiment analysis is a conceptually challenging linguistic problem as it is hard to extract sentiment polarity from sentences that depend on the speaker's context of sentiments \cite{DBLP:journals/corr/MantylaGK16}. In one of the earlier papers on sentiment analysis, \cite{DBLP:journals/corr/cs-CL-0409058} attempted to categorize positive and negative polarity from a corpus of movie reviews mined from Rotten Tomatoes. In this paper, they noted that movie reviews were particularly difficult to classify when compared to product reviews – in part because, in addition to their inherent subjectivity, words that convey negative affect are commonplace in the horror genre and should not be taken to convey overall negative sentiment \cite{DBLP:journals/corr/cs-LG-0212032}. In order to classify these reviews, they used an SVM. Similar to the objective of contrastive learning, SVM's aim to find the optimal hyperplane that distinguishes data points of each class from one another. In essence, when used to determine the polarity of movie reviews, an SVM will seek to find the boundary between the positive and negative reviews.

Contrastive learning functions in a similar way; it is a framework to distinguish polarity by organizing data into positive and negative pairs and pulling these positive pairs together while non-pairs are pushed apart \cite{1640964}. While contrastive learning techniques have seen great success when applied to computer vision problems, it was only more recently that their application has been widened to include NLP tasks such as sentiment analysis  ~\citep{rim2021adversarial,liao2021sentence, shen2021contrastive}. Contrastive learning can be a powerful addition to learning sentence polarity when used in conjunction with more traditional transformer-based models. This paper explores several methods of extracting vector representations that capture the underlying contextual and semantic features for achieving a real-world performance boost on the supervised sentiment classification tasks.

We conduct a comprehensive evaluation by fine-tuning two traditional BERT-based pre-trained models ~\citep{bert-DBLP:journals/corr/abs-1810-04805, roberta-DBLP:journals/corr/abs-1907-11692} and one supervised and unsupervised version of contrastive learning based pre-trained models ~\citep{gao-DBLP:journals/corr/abs-2104-08821} on five datasets ~\citep{potts-DBLP:journals/corr/abs-2012-15349,DBLP:journals/corr/ZhangZL15,socher-etal-2013-recursive}. We use the DynaSent datasets ~\citep{potts-DBLP:journals/corr/abs-2012-15349}for model training and report our performance on the test dataset. We achieve an average macro F1-score of 80.81\% on DynaSent-r1 and 69.08\% on DynaSent-r2 using fine-tuning of contrastive learning-based pre-trained models, an improvement of 1.32\% on DynaSent-r1 and 3.07\% on DynaSent-r2 over BERT-based fine-tuned models. We also achieve competitive performances on transfer tasks for cross-domain datasets. Finally, we consolidate the results from our experiment settings, identify class imbalance issues in the dataset and propose future work to address these challenges in evaluating sentiment analysis using text embeddings.

\section{Related Work}
The most recent sentence embeddings work focuses on solving various bottlenecks of state-of-the-art BERT-based models ~\citep{DBLP:journals/corr/abs-1908-10084,DBLP:journals/corr/abs-2103-15316}. One of the most common approaches of extraction of BERT-based embeddings for clustering and semantic search is to average the BERT output layer or use the output of the first token (the [CLS] token). However, using these practices could yield poor sentence embeddings, even worse than averaging the GloVe embeddings ~\citep{DBLP:journals/corr/abs-1908-10084}. The recent research also discusses the anisotropy problem, which hinders the BERT-based models from fully utilizing the underlying textual semantic features ~\citep{DBLP:journals/corr/abs-2103-15316}.

Much of the inspiration for this paper stems from the contrastive learning embedding work achieved by \citet{gao-DBLP:journals/corr/abs-2104-08821} which established that using contrastive learning embeddings showed improvement over various transformer-based sentence embeddings when applied to semantic textual similarity tasks. The paper presents using sentence embeddings learned in this space and comparing them to more traditional state-of-the-art embeddings such as BERT or RoBERTa \citep{bert-DBLP:journals/corr/abs-1810-04805,roberta-DBLP:journals/corr/abs-1907-11692}.Contrastive learning methods could be applied in both supervised and unsupervised machine learning settings. These methods are widely used in computer vision applications that rely on augmented versions of each input image; however, in natural language processing, it is more challenging to construct text augmentations. The recent research work on unsupervised learning approaches of contrastive learning in the domain of natural language has shown promising results. We explore techniques to learn sentence embedding in this space to achieve even better results for our sentiment analysis task.

We also draw upon the importance of dropouts in the contrastive learning architecture, as demonstrated by \citet{wu2021esimcse}. This paper suggests a modification to the unsupervised Sim-CSE (unsup-SimCSE) method for learning sentence embeddings. Specifically, the paper demonstrates that since unsup-SimCSE passes sentences through a pre-trained Transformer encoder with a dropout mask to build positive pairs, the length of each portion of the pair will be the same for the corresponding embeddings. Through experiments, they establish that ESimCSE outperformed unsup-SimCSE on both BERT-base, BERT-large, RoBERTa-base, and RoBERTa-large.

\section{Data}

This paper uses the DynaSent dataset for training and validation and tests on the DynaSent, SST-3, Yelp, and Amazon datasets~\citep{potts-DBLP:journals/corr/abs-2012-15349,DBLP:journals/corr/ZhangZL15,socher-etal-2013-recursive} (Table 1).

\begin{table*}[ht]
\centering
\begin{tabular}{p{1.5cm}p{1cm}p{1cm}p{1cm}p{1cm}p{1cm}p{1cm}p{1cm}p{1cm}p{1cm}}
& \multicolumn{3}{c}{\textbf{DynaSent r1}} & \multicolumn{3}{c}{\textbf{DynaSent r2}} & \textbf{SST-3} & \textbf{Yelp} & \textbf{Amazon} \\
& Train & Dev & Test & Train & Dev & Test & Test & Test & Test\\
\textbf{Positive} & 21391 & 1200 & 1200 & 6038 & 1200 & 240 & 909 & 10423 & 110573\\
\textbf{Negative} & 14021 & 1200 & 1200 & 4579 & 1200 & 240 & 912 & 9778 & 107967 \\
\textbf{Neutral} &  45076 & 1200 & 1200 & 2448 & 1200 & 240 & 389 & 4799 & 57810 \\
\\
\textbf{Total} & 80488 & 3600 & 3600 & 13065 & 3600 & 720 & 2210 & 25000 & 276350 \\
\end{tabular}
\caption{Datasets used for Experiments}
\end{table*} 

\subsection{DynaSent Dataset}

The DynaSent dataset \cite{potts-DBLP:journals/corr/abs-2012-15349} consists of 121,634 sentences with Positive, Neutral, and Negative labels. It is an adversarial dataset that was generated in 2 rounds. In the first round, Model 0, a RoBERTa-base model with a three-way sentiment classifer head, was trained on a sample of sentences from the Consumer Reviews \cite{consumer-reviews}, IMDB \cite{maas-imdb}, SST-3 \citep{socher-etal-2013-recursive}, and Yelp and Amazon datasets \cite{DBLP:journals/corr/ZhangZL15} to identify challenging sentences. These sentences were then human-validated and used to generate the Round 1 dataset. A second model, Model 1, was then trained on the Round 1 dataset and the 5 original datasets. Crowdworkers were prompted to use the Dynabench Platform to come up with sentences that would fool Model 1. As before, the crowdsourced dataset was human-validated to produce the Round 2 dataset. 

\subsection{Yelp and Amazon Datasets}
These datasets consist of reviews of restaurants and products collected from Yelp and Amazon and were put together by \citet{DBLP:journals/corr/ZhangZL15}. The reviews on both Yelp and Amazon originally had labels on a five-point scale, so for the purpose of our task, we re-mapped labels of 1 or 2 to Negative, 3 to Neutral, and 4 or 5 to Positive. 

The full Yelp dataset consists of 1,569,264 reviews from the Yelp Dataset Challenge 2015.

The full Amazon dataset consists of 34,686,770 reviews from 6,643,669 users on 2,441,053 products collected over a period of 18 years.

\subsection{SST-3 Dataset}
The Stanford Sentiment Treebank dataset consists of 11,855 sentences collected from movie reviews ~\citep{socher-etal-2013-recursive}. In the SST-3, sentences are labeled either Positive, Negative, or Neutral.

\section{Models}
For our model architecture (Figure 1), we use pre-trained transformer embeddings as the embedding layer with a three-way classification head on top. The classification head consists of two dropout layers, a hidden dense layer, an activation layer, and finally, a linear layer with 3-dimensional output – one for each sentiment label. We have explored various combinations of pre-trained embeddings, hidden dense layers, activation functions, and loss functions in our models, as described in the later sections.

\begin{figure}[ht]
\centering
\includegraphics[scale=0.45]{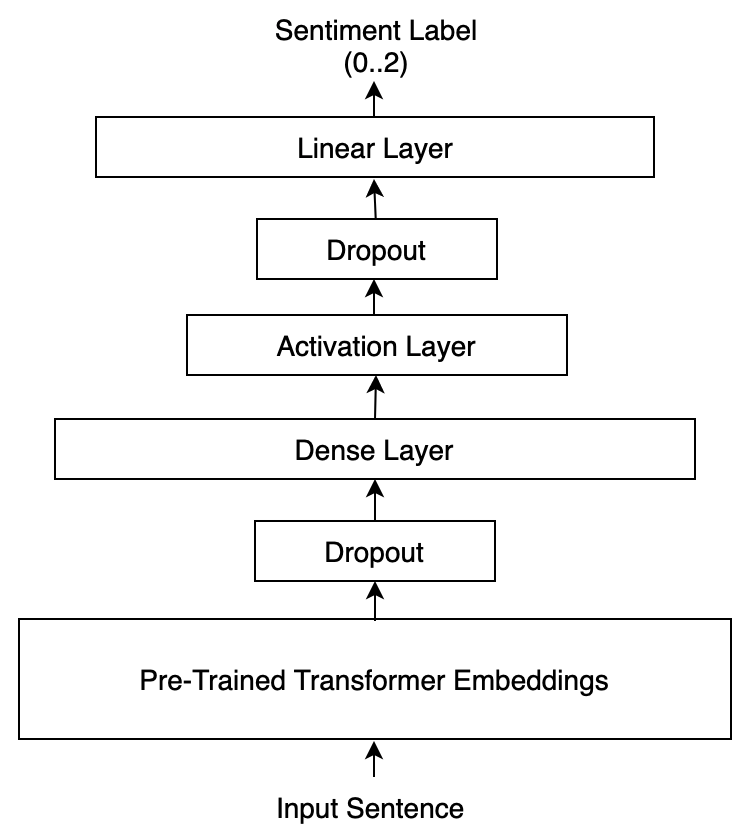}
    \caption{Sentiment Classification Model Architecture}
    \label{fig:model}
\end{figure}

\subsection{Pre-Trained Embeddings}

We use four different pre-trained transformer embeddings, all of which are available with the Hugging Face's Transformers library \cite{wolf2020huggingfaces}. The output embedding size of all these pre-trained embeddings is 768. 

\subsubsection{BERT base uncased}

BERT base uncased embeddings are pre-trained on the English language self-supervised using two objectives - masked language modeling (MLM) and Next Sentence Prediction (NSP). These embeddings are case insensitive. These embeddings were introduced in the original BERT \cite{bert-DBLP:journals/corr/abs-1810-04805} paper and serve as baseline embeddings for our models.

\subsubsection{RoBERTa base}

RoBERTa \cite{roberta-DBLP:journals/corr/abs-1907-11692} base embeddings are also pre-trained on the English language in a self-supervised fashion. Unlike BERT, these embeddings were trained on the MLM objective only. These embeddings have shown increased performance over the original BERT embeddings. Like BERT, these embeddings are case insensitive. 

\subsubsection{Sim-CSE RoBERTa base supervised}

Sim-CSE RoBERTa base supervised \citep{gao-DBLP:journals/corr/abs-2104-08821} embeddings are pre-trained on various natural language inference (NLI) datasets using RoBERTa base parameters in a contrastive learning fashion. The supervised training approach incorporates annotated pairs from natural language inference datasets into the contrastive learning framework by using "entailment" pairs as positives and "contradiction" pairs as hard negatives. 

\subsubsection{Sim-CSE RoBERTa base unsupervised}

Sim-CSE RoBERTa base unsupervised ~\citep{gao-DBLP:journals/corr/abs-2104-08821} embeddings are pre-trained on the Wikipedia corpus using RoBERTa base parameters in a contrastive learning fashion. The unsupervised training approach takes an input sentence and predicts itself in a contrastive objective, with only standard dropout used as noise. As per the original paper, this method works surprisingly well and performs on par with previously supervised counterparts. 

Both Sim-CSE RoBERTa base supervised and unsupervised embeddings have shown improvement over various NLP tasks and form the basis of our hypothesis set in this paper.

\subsection{Classification Head}

We experimented with three different dense layers for the classification head - Linear, BiGRU, and BiLSTM (Table 2). 
\begin{table}[h!]
    \centering\small
    \begin{tabular}{|c|c|c|c|c|c|}
        \hline
        \multicolumn{3}{|c|}{Dense Layer} & \multicolumn{2}{c|}{Linear Layer} \\
        \hline
        Type & In & Out & In & Out \\
        \hline
        Linear & 768 & 768 & 768 & 3 \\
        BiGRU & 768 & 256 & 512 & 3\\
        BiLSTM & 768 & 256 & 512 & 3\\
        \hline
    \end{tabular}
    \caption{Classification Head Configurations}
    \label{tab:my_label}
\end{table}
For dropout layers, the drop probability value is set to 0.1 for all the models. For our activation layer, Tanh and ReLU activation functions are used for Linear and, BiGRU and BiLSTM, respectively.

\subsection{Loss Functions}

In our experiments, we have used Cross-Entropy and Focal loss functions.

\subsubsection{Cross-Entropy Loss}

Cross-entropy loss is also known as logarithmic loss, logistic loss, or log loss. Each predicted class probability is compared to the actual class desired output, 0 or 1. A score/loss is calculated that penalizes the probability based on how far it is from the actual expected value. The penalty is logarithmic, yielding a large score for large differences close to 1 and a small score for small differences tending towards 0. The aim is to minimize the loss, i.e., the smaller the loss, the better the model. A perfect model has a cross-entropy loss of 0.

\[ Cross Entropy(p_t) = -log(p_t)\]

\subsubsection{Focal Loss}
Given the nature of imbalance in classes during training, we used the Focal Loss function \cite{lin2018focal}. Focal loss is an improvement over cross entropy loss to address imbalance issue. It applies a modulating term to the cross-entropy loss to focus learning on hard negative examples. It is a dynamically scaled cross-entropy loss, where the scaling factor decays to zero as confidence in the correct class increases. The idea is that the scaling factor will automatically down-weight the contribution of easy examples during training and rapidly focus the model on hard examples.
Focal loss adds a factor to the standard cross-entropy criterion with a tunable focusing parameter.

\[ Focal Loss(p_t) = -(1-p_t)^\gamma log(p_t), \gamma\geq0\]

\section{Experiments}

We trained our model with 93547 train samples and performed hyperparameter tuning on 4320 dev samples (Figure 2) from combined DynaSent-r1 and DynaSent-r2 datasets. However, we noticed a class imbalance between positive, negative, and neutral labeled samples.

\begin{figure}[h!]
\centering
\includegraphics[scale=0.15]{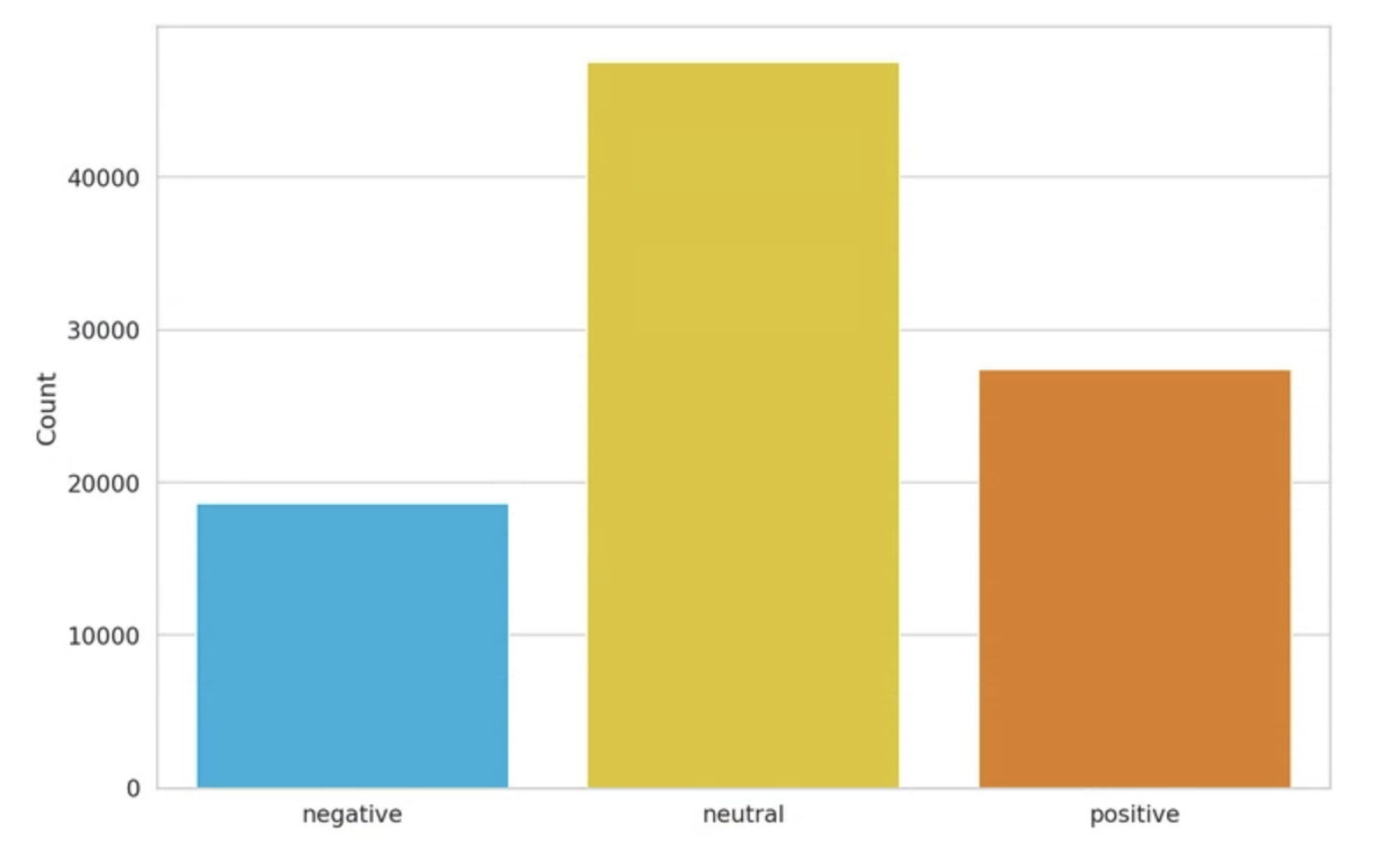}\includegraphics[scale=0.15]{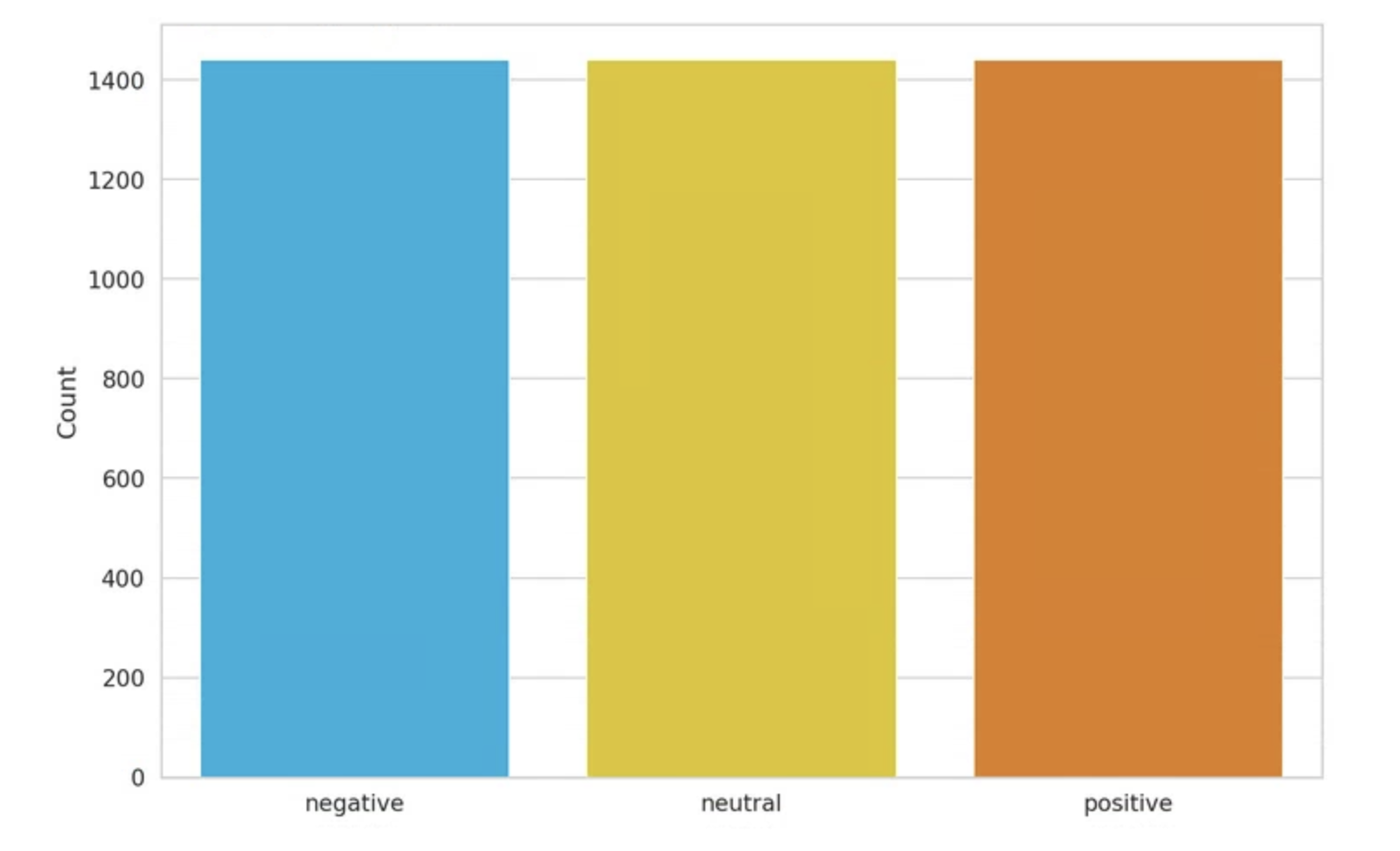}
    \caption{Sample Distribution in Train (left) and Test (right) Datasets}
    \label{fig:sample distribution}
\end{figure}

The best experiment result had a macro F1-score of 81.43\% with an accuracy of 81.5\% when evaluated on DynaSent-r1. This result was achieved using a Linear Classifier head model on a fine-tuned unsupervised SimCSE RoBERTa base pre-trained model with Cross-Entropy Loss. For DynaSent-r2, the best result had a macro F1-score of 70.46\% and accuracy of 70.56\%, which was achieved by fine-tuning on an unsupervised SimCSE RoBERTa base pre-trained model with a Linear Classification head and a Focal Loss of gamma value 3. We achieved an average macro F1-score of 80.81\% on DynaSent-r1 and 69.08\% on DynaSent-r2 using fine-tuning of contrastive learning-based pre-trained models, an improvement of 1.32\% on DynaSent-r1 and 3.07\% on DynaSent-r2 over BERT-based fine-tuned models (Figures 3, 4).

We also achieved competitive results for our transfer learning evaluations on cross-domain datasets (Appendix Figure 5). The best experiment result had a macro F1-score of 63.3\% with an accuracy of 70\% when evaluated on SST3.This result was achieved using a BiLSTM Classifier head model on a fine-tuned unsupervised SimCSE RoBERTa base pre-trained model with Focal Loss.For the Amazon dataset, the best result had a macro F1-score of 52.53\%, which was achieved by fine-tuning on a supervised SimCSE RoBERTa base pre-trained model with Focal Loss. For the Amazon dataset, the best result had a macro F1-score of 52.53\% and accuracy of 54.33\%, which 
\clearpage
\begin{figure}[h!]
    \captionsetup{justification=centering}
    \includegraphics[scale=0.55]{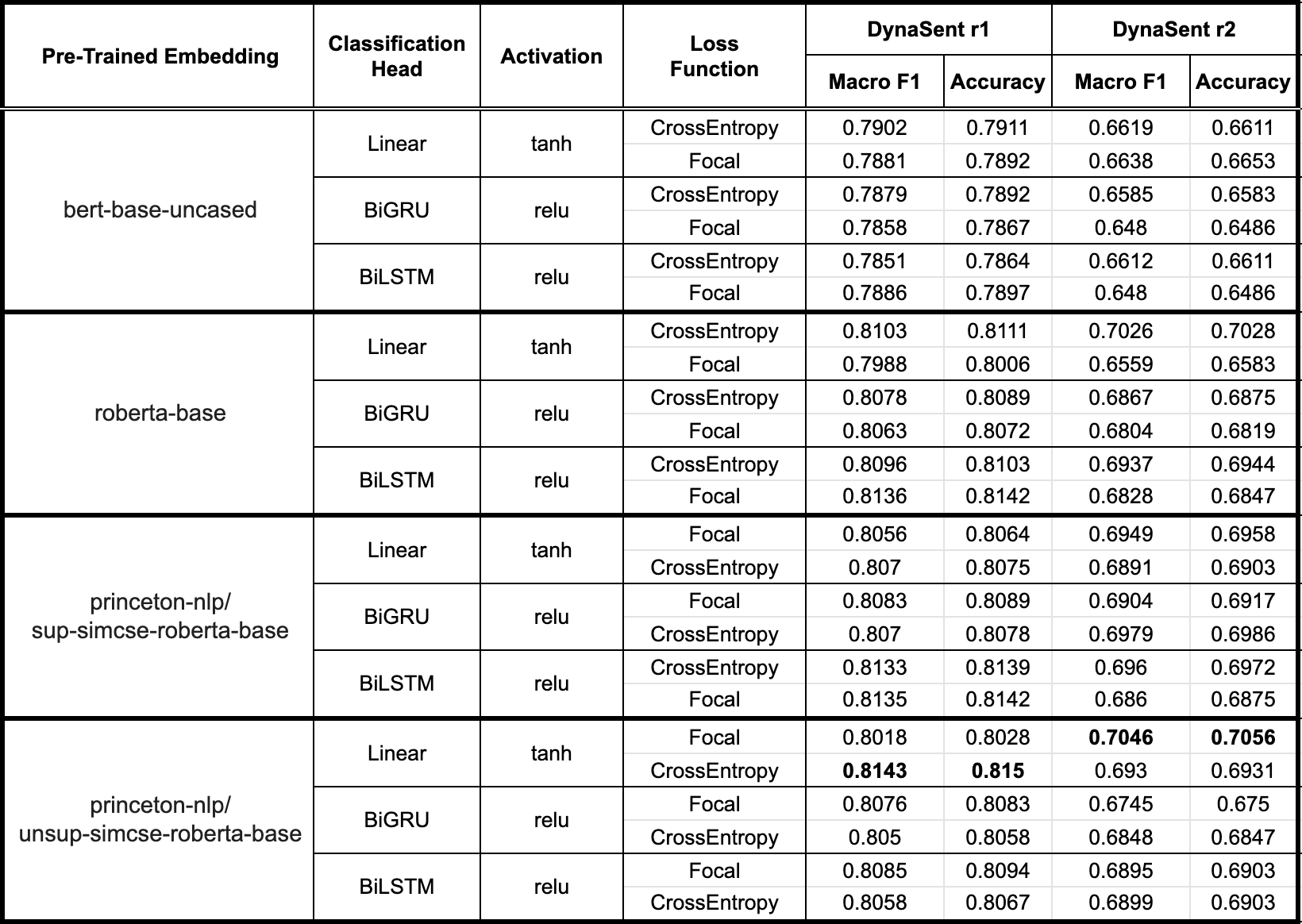}
    \caption{Experiment Results for DynaSent Datasets.} 
    \label{fig:dynasentdata1}
    \includegraphics[scale=0.48]{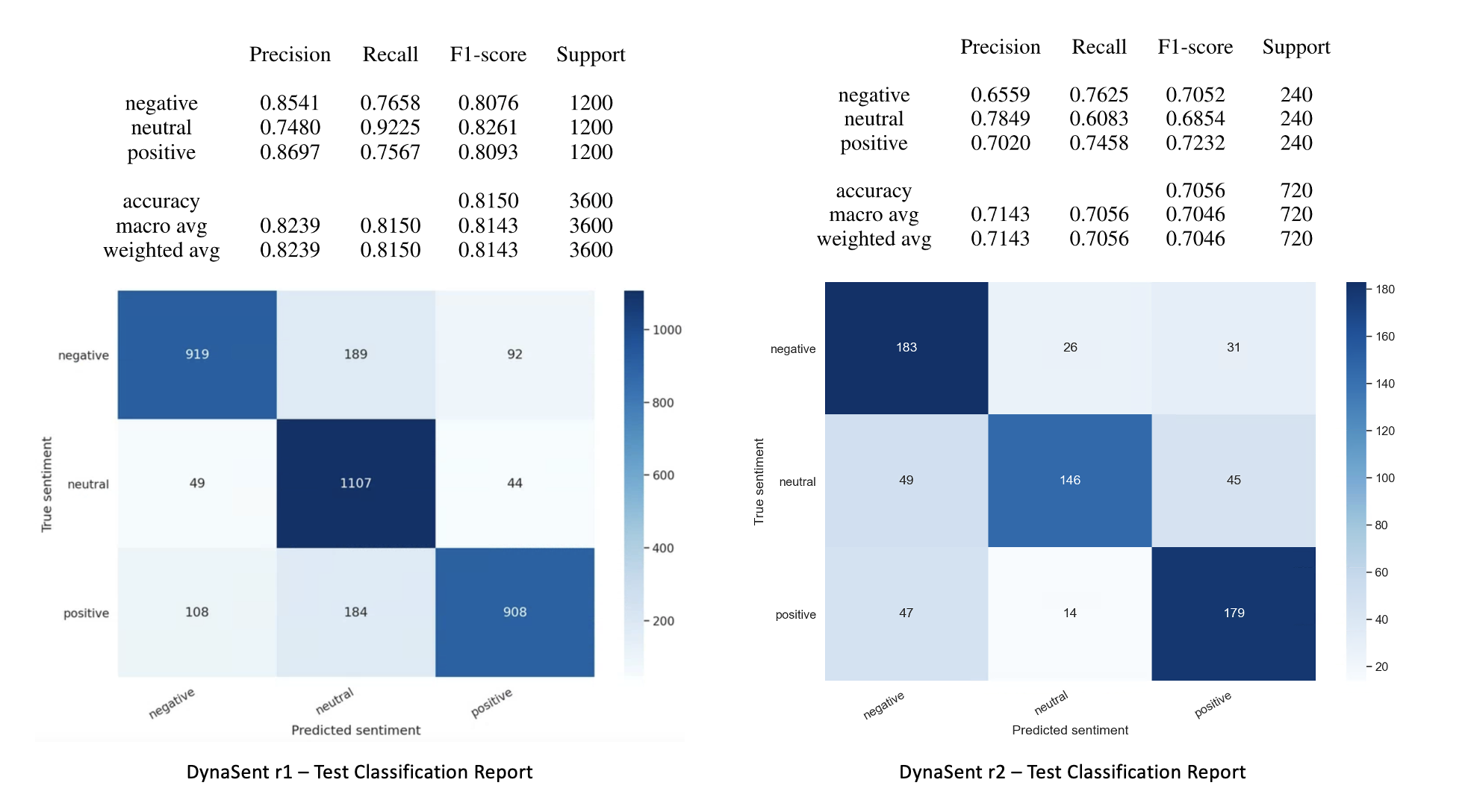}
    \caption{Best Results for DynaSent Datasets}
    \label{fig:classification report3}
\end{figure}
\clearpage

was achieved by fine-tuning on a supervised SimCSE RoBERTa base pre-trained model with a BiLSTM Classification with Cross-Entropy Loss.However, for the Yelp dataset, the traditional BERT-based models outperformed the contrastive learning-based ones.

For our experiments, we performed hyperparameter tuning across activation functions, dropouts, loss functions, and hidden sizes for various classification heads (Table 3). Our experiments concluded that Tanh activation performs better with a Linear classification head and ReLU with BiGRU and BiLSTM classification heads. The max length was set to 64 for all test sets evaluation. However, increasing the max length to 256 from 64 increased the accuracy by around 6\% for the Yelp dataset. 

We limited our experiments to use two BERT-based pre-trained embeddings (e.g., bert-base- uncased and roberta-base) and observed that fine-tuning performed on contrastive learning-based embeddings (e.g., supervised unsupervised simcse-roberta-base) consistently outperformed its traditional counterparts. 
\renewcommand{\arraystretch}{1}
\begin{table}[h!]
    \centering\small
    \begin{tabular}{|c|c|}
        \hline
        Hyper-Params & Values \\
        \hline
        Dropout & 0.1 \\
        Activation & Tanh, ReLU \\
        Focal Loss-gamma & 3 \\
        Focal Loss-reduction & mean \\
        Max-Length & 64 \\
        Batch-Size & 32 \\
        Optimizer & AdamW \\
        Learning Rate & 1e-5 \\ 
        Weight Decay & 0.01 \\
        Epochs & 4 \\
        \hline
    \end{tabular}
    \caption{Hyper-parameters Tuned for Experiments. (Activation) Tanh for Linear Classification Head and ReLU for BiGRU and BiLSTM Classification Heads. (Optimizer) PyTorch 1.10 implementation of AdamW}
    \label{tab:hyperparams}
\end{table}

\section{Analysis}
Based on our experiments, we observed that contrastive learning-based models significantly outperformed the traditional BERT-based models. The performance boost over the traditional BERT-based pre-trained model is because of the key properties of contrastive loss used in the pre-trained contrastive models, which optimizes on the alignment of features from positive pairs, and uniformity of induced distribution of the normalized features. Also, instead of using direct text representations from pre-trained models, we focused on fine-tuning models with additional neural network layers to derive sentiment-based text embeddings for our task evaluation. We also observed that using this approach, we were able to classify better the sentiments for movies reviews, particularly the horror movie reviews, containing more negative words but not necessarily conveying a negative sentiment.

We also observed that the best model trained on the DynaSent-r1 and DynaSent-r2 datasets did not perform as expected when evaluated on cross-domain test datasets, particularly on the restaurant reviews for the Yelp dataset. Also, the decrease in performance was especially true for the Neutral classes across all cross-domain datasets. The original DynaSent paper \cite{potts-DBLP:journals/corr/abs-2012-15349} described the Neutral class as being tricky to evaluate for a few different reasons, so, unsurprisingly, the performance for this class would be diminished. Although we used Focal Loss in our experiments to implicitly address the class imbalance issues, we need to address this issue explicitly, and the following section lists a few approaches.

\subsection{Future Work}
\setlength{\parskip}{0pt}
Particularly for the SST-3, Amazon, and Yelp datasets, the number of samples per class is largely imbalanced. In addition to being an underrepresented class, a three-star review could convey vastly differing semantic meanings \cite{potts-DBLP:journals/corr/abs-2012-15349}. For the Amazon and Yelp datasets, it could indicate ambivalence or mixed-sentiment towards the product or place that was being reviewed; for SST-3, it could simply indicate that the reviewer was unsure of the author's underlying sentiment. In either case, the performance of our model on the underrepresented Neutral class when evaluating on these datasets is not as optimal. In order to rectify this, we explored using upsampling techniques to generate more examples of the Neutral class so that we could train on this aggrandize dataset. 

One upsampling technique of interest was presented by \citet{Scott2019GANSMOTEAG}. In this paper, \citet{Scott2019GANSMOTEAG} merges generative adversarial networks (GANs) \cite{goodfellow2014generative} with Synthetic Minority Oversampling Technique (SMOTE) \cite{smote-2002}, a statistical technique that in its raw form doesn’t map well to text data. Using the GAN-SMOTE architecture, they were able to generate convincing synthetic examples of their one-hot encoded dataset and improve the accuracy of their experiment benchmark. 

While the GAN-SMOTE architecture was interesting, another, simpler method of upsampling was explored as part of this paper using the Fill Mask Pipeline offered through HuggingFace's Transformers library \cite{wolf2020huggingfaces}. Specifically, we experimented with using the Fill Mask model to generate synthetic sentences. Fill Mask works by replacing a word in the input sentence with a semantically similar one. After training on a subset of these upsampled Neutral sentences though, it became apparent that the model was overfitting as the training and validation accuracy scores were vastly different. Additional work to correct this issue can be done in a future iteration of this project. 

\section{Conclusion}
In this work, we propose a contrastive learning-based fine-tuning approach, which outperforms state-of-the-art BERT-based text embeddings for our sentiment analysis task. We fine-tune on both the unsupervised pre-trained model, which predicts input sentiments with dropout as noise for data augmentation, and the supervised pre-trained model, which originally uses NLI datasets for pre-training. Through our experiments, we extracted superior sentiment-based text embeddings learned in a contrastive learning space to achieve better results when evaluating on DynaSent datasets. Additionally, we analyze the test results for sentiments classification using transfer learning on cross-domain datasets. We also propose our future work to address class imbalance issues, which will help improve upon the benchmark results.

\section{Acknowledgement}
The authors would like to thank Dr. Christopher Potts, Professor, and Chair of Linguistics, and, Professor of Department of Computer Science, Stanford University, California, USA, for all his support and guidance.

\bibliography{acl2020}
\bibliographystyle{acl_natbib}

\clearpage
\appendix
\section{Appendix}
\begin{figure}[h!]
    \centering
    \includegraphics[scale=0.45]{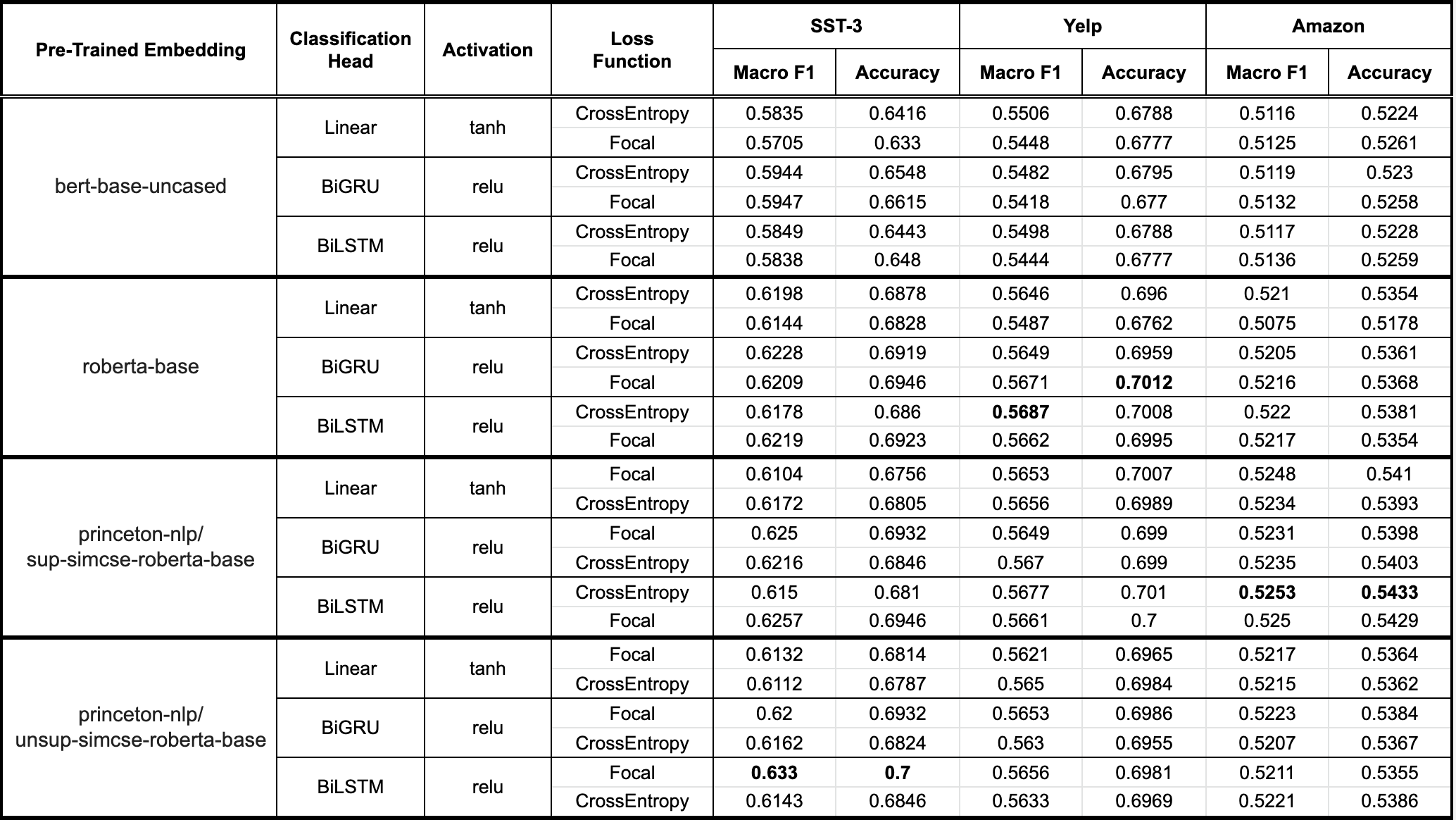}
    \caption{Experiment Results for Other Datasets.}
    \label{fig:otherdata2}
\end{figure}

\end{document}